\documentclass[sigconf,preprint]{acmart}

\usepackage{CJKutf8}
\usepackage{tcolorbox}
\tcbuselibrary{breakable}

\AtBeginDocument{%
  }

\usepackage{array}
\usepackage{tabularx}
\usepackage{ragged2e}
\newcolumntype{S}{>{\RaggedRight\ttfamily\arraybackslash}X}
\newcolumntype{L}{>{\RaggedRight\arraybackslash}X} 
\setcopyright{rightsretained}
\copyrightyear{2025}
\acmYear{2025}
\acmConference[Genaiecom '25]{Proceedings of the second workshop on Generative AI for E-Commerce}{September 22, 2025}{Prague, CZ}
\acmBooktitle{Proceedings of the second workshop on Generative AI for E-Commerce 2025, September 22, 2025}
\acmYear{2025}
\acmMonth{09}
\acmDOI{}
\acmISBN{}
\acmPrice{15.00}

\begin{document}

\title{Using item recommendations and LLMs in marketing email titles}

\author{Deddy Jobson}
\email{deddy@mercari.com}
\affiliation{%
  \institution{Mercari Inc.}
  \city{Tokyo}
  \country{Japan}
}

\author{Muktti Shukla}
\email{m-shukla@mercari.com}
\affiliation{%
  \institution{Mercari Inc.}
  \city{Tokyo}
  \country{Japan}
}

\author{Phuong Dinh}
\email{pdinh@mercari.com}
\affiliation{%
  \institution{Mercari Inc.}
  \city{Tokyo}
  \country{Japan}
}

\author{Julio Christian Young}
\email{jc.young@mercari.com}
\affiliation{%
  \institution{Mercari Inc.}
  \city{Tokyo}
  \country{Japan}
}

\author{Nick Pittoni}
\email{nick.p@mercari.com}
\affiliation{%
  \institution{Mercari Inc.}
  \city{Tokyo}
  \country{Japan}
}

\author{Nina Chen}
\email{nina.chen@mercari.com}
\affiliation{%
  \institution{Mercari Inc.}
  \city{Tokyo}
  \country{Japan}
}

\author{Ryan Ginstrom}
\email{r-ginstrom@mercari.com}
\affiliation{%
  \institution{Mercari Inc.}
  \city{Tokyo}
  \country{Japan}
}
\renewcommand{\shortauthors}{Jobson et al.}

\begin{abstract}
  E-commerce marketplaces make use of a number of marketing channels like emails, push notifications, etc. to reach their users and stimulate purchases. Personalized emails especially are a popular touch point for marketers to inform users of latest items in stock, especially for those who stopped visiting the marketplace. Such emails contain personalized recommendations tailored to each user's interests, enticing users to buy relevant items. A common limitation of these emails is that the primary entry point, the title of the email, tends to follow fixed templates, failing to inspire enough interest in the contents. In this work, we explore the potential of large language models (LLMs) for generating thematic titles that reflect the personalized content of the emails. We perform offline simulations and conduct online experiments on the order of millions of users, finding our techniques useful in improving the engagement between customers and our emails. We highlight key findings and learnings as we productionize the safe and automated generation of email titles for millions of users. 
\end{abstract}

\keywords{Email Title, Generative AI, large language models, recommendations, TnS}

\received{10 August 2025}
\received[accepted]{28 August 2025}

\maketitle
\begin{CJK*}{UTF8}{gbsn}
\section{Introduction}
In order to stay relevant in the minds of consumers, companies run a plethora of marketing initiatives spending trillions of yen annually in Japan\cite{noauthor_2023_2024}. Marketing initiatives can be decomposed into the target, the message, and the distribution channel\cite{mccarthy_e_jerome_and_perreault_william_d_basic_1987}. For a large C2C marketplace like Mercari, with tens of millions of active users, in-app touch points such as in-app notifications, banners, etc. are strong touchpoints to convey marketing messages. However, to reach users who stopped participating in the marketplace (and therefore do not log into the service anymore), external touch points like search ads, television commercials, etc. are more effective. 

Emails are especially suited as a method to reach customers who once used Mercari but stopped, since it allows us to deliver rich content personalized with recommendations\cite{korayem_macro-optimization_2016,sharma_enhancing_2022,grbovic_e-commerce_2015} of items they are potentially interested in, having a stronger impact on the user's interest in coming back to Mercari. 

While many of the above-cited studies have investigated how to improve the content of emails, there is a dearth of research that explore how the title of marketing emails can be optimized to improve the open-rate of the marketing email. One potential reason could be the difficulty in automatically generating email titles that conform to safety standards with minimal risk. With the rise of adoption of generative AI in marketing\cite{cillo_generative_2025}, the reliability of generated titles has increased to the point of being trustworthy with limited human oversight. In this paper, we conduct an experiment in Mercari where we distribute emails with personalized titles improved by large language models\cite{xiao_foundations_2025}. Our contributions are as follows:
\begin{itemize}
    \item We demonstrate the value added by using large language models to generate personalized titles in marketing emails through large scale experiments with over a million users. 
    \item We explain how we performed multiple levels of quality assurance checks and how we iterated over such hurdles to deliver AI-generated email titles responsibly.
    \item We gather our findings to develop a framework to ensure such customer-facing generated titles are safe and reflect Mercari's brand image. 
\end{itemize}

\section{Research Questions}
To make clear our objectives, we investigate the following research questions: 
\begin{itemize}
    \item \textbf{RQ1:} Can email titles be improved through LLMs? This can be verified through changes in the open rate. 
    \item \textbf{RQ2:} Can the improvement in emails offset the costs of using LLMs? This would depend on the value associated with the opening of an email (connected through downstream actions like purchases).
\end{itemize}

\section{Development}

\subsection{Open Source vs Proprietary Models}
When considering which large language model to deploy, we consider two options: locally deploying open source models, and calling proprietary models through APIs. Generating titles through open source models give us more control over the version of the model used and also open up the possibility of finetuning in the future while potentially being cheaper to run. Proprietary models can give us higher quality titles at the cost of reduced flexibility and increased API costs. 

We performed offline simulations with two candidates models:
\begin{itemize}
    \item \textbf{Open Source: }Llama 3.2 - 3 billion parameter version
    \item \textbf{Proprietary: }GPT 4o-mini
\end{itemize}

We choose the 3 billion parameter version since it can generate titles quickly without requiring dedicated GPUs making it very cost effective. However, as seen in Table \ref{tab:llama_vs_4o_mini_outputs}, we found the local model's generated titles rather bland. Furthermore, we find proprietary models to better adhere to the structured outputs and other requirements to be production ready. Both disadvantages can be attributed to the small size of the model. With newer advances in small open weights LLMs for solving specialized problems\cite{gunasekar_textbooks_2023}, we expect this gap to be bridged in the long run. For our experiments however, the estimated cost of using OpenAI's API was around 300 USD per week, which was well within the budget for a proof-of-concept (POC) experiment. We therefore proceed with the proprietary model. 

\begin{table*}[t]                
  \centering
  \footnotesize                
  \begin{tabularx}{\textwidth}{|@{}S|S@{}|} 
    \hline
    \textbf{Titles of recommended items inside email} & \textbf{Generated Title} \\\hline

    "Apple iPhone Case", "Wireless Mouse", "Bluetooth Speaker" &
    Elevate Your Mobile \& Connect: iPhone Case, Wireless Mouse \&
    Bluetooth Speaker Bundle \\[.4ex]
    \hline
    "Insulated Water Bottle", "Ergonomic Office Chair",
    "Noise-Cancelling Headphones", "LED Desk Lamp" &
    Boost Your Productivity: Insulated Water Bottle, Ergonomic Chair
    \& More Essential Accessories \\[.4ex]
    \hline
    "Gaming Keyboard", "Portable Charger", "Smart Watch",
    "Fitness Tracker", "Window Curtains", "Cookware Set",
    "Comfortable Pillow", "Bluetooth Earbuds", "Laptop Stand",
    "Dry-Erase Board" &
    Revamp Your Routine: Gaming Keyboard, Smart Watch, Laptop Stand
    \& More Essentials \\
    \hline
  \end{tabularx}
  \caption{Samples of email titles generated by a local open-source model (Llama 3.2-3b)}
  \label{tab:llama_vs_4o_mini_outputs}
\end{table*}

\begin{figure}[htbp]
\centering
\includegraphics[width=0.8\linewidth]{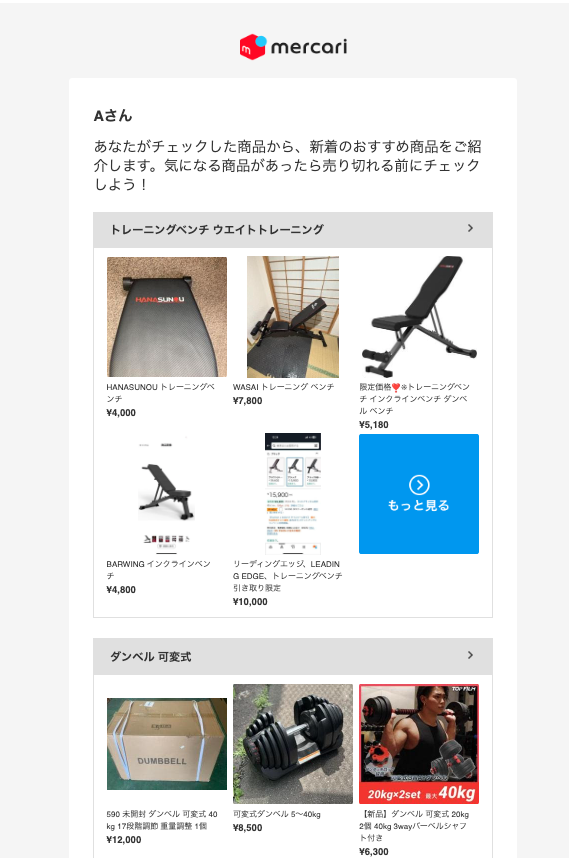}
\caption{Sample Email with Item Recommendations}
\Description{A screenshot of a marketing email on a mobile phone. The email contains several recommended items for sale, including a camera, a backpack, and headphones, each with an image and price.}
\label{fig:sample}
\end{figure}

\begin{figure}[htbp]
\centering
\includegraphics[width=0.8\linewidth]{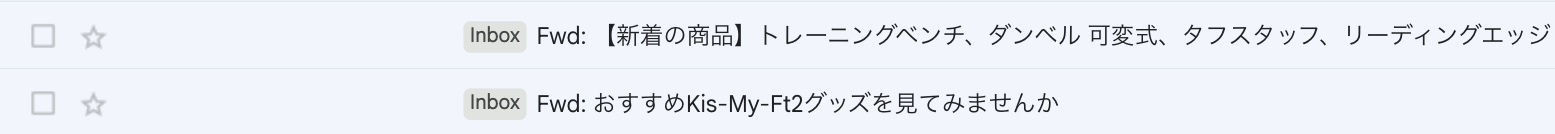}
\caption{Sample Email Titles before and after gen AI}
\Description{Two examples of email subject lines. The upper one is a template-based title: "'Nike Sneakers 28cm' and other items...". The lower one is a thematic, AI-generated title: "For your daily commute! Bags, sneakers, and more."}
\label{fig:sample_titles}
\end{figure}

\subsection{Context Engineering}
After selecting OpenAI's GPT 4o-mini as the model for this experiment, we create a prompt to guide LLM on how to generate an email title for each user based on their search log. Our prompt engineering aims to clarify 5 main rules: technical parameters, content structure, tone and style, call-to-action (CTA) guideline, and prohibited wording. The content structure is to ensure the generated title is relevant to the email contents, the tone and style defines a professional-yet-approachable brand voice and the language is tailored to the specific audience, the CTA section is to restrict each email to one among a group of specific CTA, and lastly the prohibited wording consists of a set of rules in order to comply with company's policy. Aligning the brand voice is especially a challenge to achieve in Japanese, where the tone can shift dramatically from nuanced changes in formality\cite{ide_formal_1989}. We enlisted the help of native speakers of Japanese to ensure the CTAs expressed our intents. The final version of the prompt gets updated after a thorough legal check. To enhance the model's performance and enable in-context learning, the prompt also included few-shot examples from historical emails sent by Mercari in the past. This method provides contrastive illustrations of both effective and ineffective subject lines, allowing the model to better generalize from the instructions. The complete prompt is available in the appendix.

\subsection{Human-in-the-loop validation}
Before sending AI-titled emails to end users, it is crucial to ensure that our system does not generate any undesirable email titles. To achieve this, our cross functional project team, consisting of project managers, engineers, and brand executives, reviewed three iterations of generated content based on a sample set of user data. We use a sample of 1,000 search keywords and item names from historical marketing emails to generate 1,000 corresponding email titles, which then underwent a thorough manual quality control process.

During this process, we screen for quality and security issues and identify several recurring problems: repetitive phrasing, awkward combinations of item names, excessive length, unnatural Japanese language, incomplete words, and the inclusion of sensitive items. These findings highlight common pitfalls in LLM-driven marketing; therefore, any team working on similar applications should anticipate and screen for these types of errors. We have some suggestions resolving these aforementioned issues. For instance, we debated whether to provide the model with single or multiple items at once and considered restricting items to the same category to improve relevance. Internal feedback also highlighted the need for greater lexical variety and subtle personalization to make the titles more appealing. To address these issues, our engineers have since refined the prompts and implemented a sensitive word filter.

\subsection{Implementation}
\begin{figure*}[t]
\centering
\includegraphics[scale=0.4]{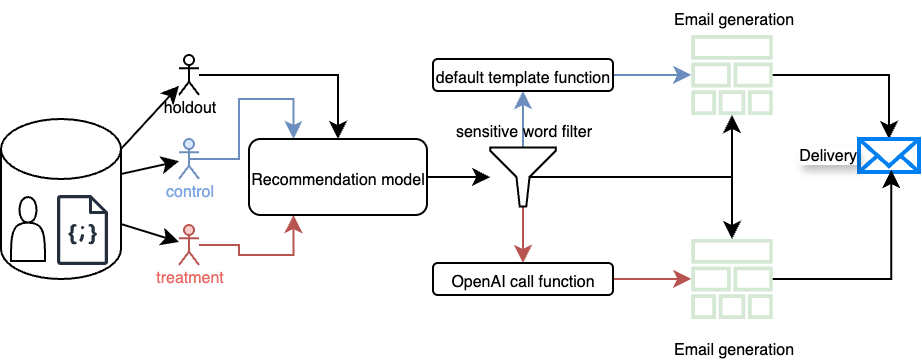}
\caption{Implementation schematic}
\Description{A flowchart diagram illustrating the technical implementation. It shows the process from User Selection, through Item Recommendation, Sensitive Word Filter, Email Title Generation (with branches for Control and Treatment groups), and finally to Email Construction and Delivery.}
\label{fig:impl}
\end{figure*}
We construct and distribute email following the flow of this schematic Figure \ref{fig:impl}.

\subsubsection{User selection and experiment assignment}
Based on users' activity, we filter users who have not accessed the Mercari app in the last 7 days and have accessed Mercari at least once in the last 1 year but made at least one purchase in the last 6 months. We then implement user-level randomization for A/B testing. The assignment mechanism uses deterministic hashing based on user identifiers to ensure consistent treatment assignment across sessions. There are two variants in this projects: treatment with LLM-generated email title's email, and control with standard recommendation email provided by our company. Details and explanations can be found in Section \ref{sec:exp_setup}.
\subsubsection{Items recommendation}
For each user, we have a set of recommendation items from the in-house tuned recommendation model. We prioritize relevant items and on-trend items for users. 
\subsubsection{Sensitive word filter}
After that, we filter out any item whose name consists of sensitive words. The check is done by comparing with a predefined weekly-updated list of sensitive words, for example in Table \ref{tab:sensitive_email_examples}.
\subsubsection{Email title generation}
The system builds a mapping that aggregates item names by category to create contextual input for the LLM in \textbf{treatment} case or top rated item names in \textbf{control} case. Based on the assignment of user's experiment, the system maintains two parallel execution paths: template-based path for \textbf{control} group and LLM-based path for \textbf{treatment} group. With template-based path, we will use a pre-defined email title structure and with LLM-based path, we will use the aforementioned metadata and invoke the external language model's API through the JSON format. Retry mechanism is implemented here with a low number of allowed retry in order to balance with the cost and in-time email delivery demand.
\subsubsection{Email component construction and delivery}
We will aggregate email title, chosen templates, and recommendation items altogether and delivery to each user (refer figure \ref{fig:sample}). A log is implemented with necessary metrics to track the progress and impact of email.

\subsubsection{LLM-as-a-Judge}
Before passing the prompt and generated outputs for evaluation to various other teams for verification, we use larger LLMs like Claude 3.5 and Gemini to evaluate the generated titles for "appropriateness". While our requirements at this stage were not precise, it was sufficient to get instant feedback for some major issues like family friendliness of the content, which we incorporate into the updated prompt.

\subsection{Legal Process} 
We conduct a head-to-toe process to release a customer-facing AI-generated application. In essence, the legal process provides necessary guardrails to ensure that while utilising AI application in company's products, we are still protecting our customers, our brand, and the company from significant legal and financial risks.
\subsubsection{Trust and Safety (TnS) Review} 
The generated email titles were guaranteed to avoid prompting inappropriate content by avoid blocked message in a pre-defined list of expletive words from TnS team. We then leverage morphological analysis filters to expel those words. We ensure adherence to the standards of Mercari to avoid promoting inappropriate content that does not align with our brand image.
\subsubsection{Marketing Team Review}
Marketing Team are able to confirm the alignment of LLM-generated content with marketing strategy and campaign objectives. This includes making sure we are not recommending items deemed sensitive or price-inflated (like rice) in Japan.
\subsubsection{Branding/UX Review}
We ask Branding/UX representative to double check the email title generated by LLM to validate visual coherence and adherence to brand voice and user experience standards. The review covers different aspects of the title's first impression such as wording, lexicographical order, and text location.
\subsubsection{Legal Review}
In this Legal Review, we assess compliance with legal guidelines and regulations related to user communication. Following the company guideline, we make sure the prompt effectively mitigates the risk of outputs containing misinformation, discrimination, bias, or harmful information to a certain extent. The privacy evaluated from using users' past search keywords is deemed low-risk. To make sure prohibited elements will be banned and UX/Branding team feedback are well-incorporated, we utilise the same title templates in this first iteration of the project. The final review is to check the prompts being well-crafted and based on samples, the likelihood of inappropriate content appearing in the output is low.
\subsubsection{Security Review}
The purpose of this check is to verify the security posture of the solution for sensitive information filtering, ensuring protection against vulnerabilities. Security team conclude that the information used to generate email title low risk in terms of security, 
\subsubsection{Intellectual Property (IP) Specification Review}
Since this email is generated by LLM, we need to ensure there will be no infringement or violation of intellectual property guidelines. We add a clarification that this email title is generated by AI/LLM at the end of every email sent in this project.
\subsubsection{AI/LLM Ethics Team Review}
The AI/LLM Ethics team evaluated ethical considerations in AI model deployment and content generation, making sure there is no bias against ethics standards. While we expect proprietary models by themselves to not output misinformation, discrimination, bias, or harmful information, we made changes to the prompt to further reinforce ethical outputs.

\begin{table*}[t]
    \centering
    \begin{tabular}{|l|c|c|}
        \hline
        \textbf{Metric} & \textbf{Relative Lift} & \textbf{z-value} \\
        \hline
        Email Send Rate & -0.90\% & -12.11 \\
        Email Open Rate & +0.46\% & 1.76 \\
        Email Item Tap Rate & +23.63\% & 8.11 \\
        Buyer Conversion Rate (via email) & 45.12\% & 1.29 \\
        Buyer Conversion Rate (overall) & -0.42\% & -0.09 \\
        \hline
    \end{tabular}
    \caption{Performance of LLM-generated titled emails}
    \label{tab:performance_ab_test}
\end{table*}

\section{Experimental Procedure}
\subsection{Setup and Execution}\label{sec:exp_setup}
With all the checks passed and the backend system implemented, we conduct an online experiment to measure the incremental business impact of LLM-generated email titles. We do so through a randomized control trial (RCT) or in other words an A/B test. We specifically target users who recently stopped accessing the Mercari app, since these users still actively engage with the emails we send. We send them emails containing relevant item recommendations can increase usage to generate a substantial business impact. We sample over a million such users and divide them into equally-sized treatment and control groups. 

The control group of users receive emails titled with a fixed template which is constructed as follows. We extract the title of the first item recommended in the email (for example, Nike Sneakers 28cm) to create a title in the following form: ``\textit{`Nike Sneakers 28cm' and other items are currently on sale right now}''. This way, through the personalized recommendations in the body of the email, we add some level of personalization to the title of the emails. Note that the original title is in Japanese.

The treatment group receives LLM-generated email titles. While the fine details are described in depth in prior sections, in short, we use LLMs to generate thematic titles based on the content of the emails. This prevents the titles from looking stale as is potentially the case in the control group's setting. 

We run the experiment for one week, with one email being sent to each user each week. This allows us to measure the repeat open rate, which we expect to especially improve from the wider variety of email titles created by the LLM. We measure core business metrics such as average buyer conversion rate (\# buyers / \# target users) and average number of transactions per group. We also measure email-specific metrics like the email send rate, email open rate, the click through rate of items in the emails, the email unsubscription rate, etc.

\begin{table*}[t]                              
  \centering
  \footnotesize
  \begin{tabularx}{\textwidth}{|@{}L |L| l |l |L@{}|}
  \hline
    \textbf{Email title} &
    \textbf{Email title - English} &
    \textbf{NG word} &
    \textbf{NG word - English} &
    \textbf{Why considered sensitive?} \\\hline

    新米コシヒカリ30kgを早い者勝ちで！ &
    First come, first served: 30kg of new Koshihikari rice! &
    30kg & 30kg &
    Very heavy item without context. May deter older users. Needs audience segmentation. \\[.4ex]
\hline
    あほの坂田グッズをチェックしよう &
    Check out Aho no Sakata goods &
    あほ & Idiot &
    Considered rude and disrespectful; could offend users. \\[.4ex]
\hline
    おまえをオタクにしてやるから全巻セットを集めました &
    We have collected the complete set of “I’ll Make You an Otaku” &
    おまえ & You (rude) &
    Informal and condescending; can sound aggressive. \\[.4ex]
\hline
    介護用シルバーカーを集めました &
    We’ve collected our silver carts for caregivers &
    バーカ & Stupid (slang) &
    Rude and mocking tone; disrespectful. \\
    \hline
  \end{tabularx}

  \caption{Sensitive-word screening examples for Japanese email titles}
  \label{tab:sensitive_email_examples}
\end{table*}

\section{Observations}

From our A/B tests, we find that most of our target metrics were not statistically significant. This is because the estimated effect size was not as large as we had hoped. Even so, the findings hint (we define "hinting coefficients" as those with $|z\text{-value}| > 1$) towards the positive effects of our new emails. The email open rate showed a positive trend (+0.46\%), though it did not reach statistical significance. 

More interestingly so, the click-through rate of items within the emails is statistically significant with a strong relative lift of 24\% (as seen in Table \ref{tab:performance_ab_test}), even though the content of the emails was generated using the same traditional recommendation algorithm. We hypothesize that the LLM-generated titles gave users a much clearer idea of what to expect inside the emails, boosting engagement. One way to test this hypothesis can be to estimate the correlation between the open rate and some relevance measure of the email title string to the body of the email. 

However, we note that there was no significant lift in the overall buyer conversion rate among targeted users. Therefore, while the changes in the emails strongly boost engagement with users, they haven't done so enough to lift the harder-to-move business metrics.

\section{Related Work}
There has been much research around methods for optimizing marketing text. Traditionally, statistical analysis and feature-heavy machine learning methods were widely used in the optimization of email subject lines. For example, \citet{emotionandvirality} employed analysis to find the effect of emotional sentiment in arousing audiences' reaction.

With the speedy development of LLMs, marketing optimization now integrates them into recommender systems. A survey by \citet{vats2024exploringimpactlargelanguage} has shown that LLMs can enhance transparency and interactivity in such systems. \citet{liu2025llmscustomizedmarketingcontent} marked a recent shift by using a retrieval-augmented system to generate keyword-specific ad copy, achieving a 9\% higher CTR in A/B tests compared to templates.

Another concrete application of LLMs in optimizing marketing text is the ability to generate email subjects. Being the gateway to a large channel for outreach, email subjects have been subject to multiple analyses~\cite{miller_psychological_2016,paulo_leveraging_2022} in a bid to improve the open rate of emails, with some even analyzing the effect of using no title at all~\cite{sappleton_email_2016}. 
Even before the advent of large language models, machine learning methods like neural networks have been used to generate titles. The work by \citet{zhang_this_2019} showed that titles generated by their model were preferable to those created by humans. 

In the paper by \citet{emailsubjectsgeneration}, authors compare different fine-tuned LLM models in generating email subjects. Note that they used the title of a single product in their prompt to generate titles. Our methodology used multiple product titles, which gave the LLMs a more difficult task of identifying an appropriate theme for the title. Furthermore, we operated on the scale of millions of users, which introduced challenges on evaluation not covered by other studies. 

Evaluating LLMs posed another challenge for us. Unlike in the case of coding, where it is easy to verify if code does not work~\cite{edirisinghe_quality_2024}, evaluating natural language outputs using LLM-as-a-Judge frameworks~\cite{gu_survey_2025} requires working with heuristic measures which are not guaranteed to work. Reducing the human load for evaluation is an avenue of future research.

\section{Future Work}

While the empirical results so far have been encouraging, they have not been definitive. Therefore, for starters, we would like to conduct the experiment again with methods to boost the power of our experiment~\cite{deng_improving_2013,jobson_covariate_2024} to validate the findings. We would also like to use causal analysis to better understand the mechanism through which LLM-generated titles better engage with users. 

Furthermore, the current prompts only take into context the items recommended in the email. In future iterations, we plan to include user-profile attributes like their historical categories of interest, demographic information, etc. to check if improved personalization of email titles drives more engagement. 

Lastly, our findings for the improved performance of email titles open the door to improving other components of our marketing emails, such as the headings of various components within the emails, personalized messaging in the body of the emails, etc. 

\section{Conclusion}
In this paper, we explore the usage of large language models for generating email subject titles. We compare locally deploying open-source models with calling proprietary models from service providers, finding proprietary models to perform better. Our online experiments on over a million users validate the utility of generative AI for marketing content generation, and open the door to optimizing other components of the email like message, sub-headings, etc. Since exposing the output of LLMs directly to end users carry risks, we perform multiple rounds of evaluation with multiple stakeholders to ensure compliance with law, safety, ethics, and brand image. The lessons learned from our end to end experience deploying llms responsibly serve as a reference for others who would like to do the same.

\begin{acks}
We would like to thank Masaki Chotoku for helping with performing the NG word assessment and also Mercari Inc. for supporting this research and providing the platform for experimentation.
\end{acks}

\bibliographystyle{ACM-Reference-Format}
\bibliography{references-deddy,references-phuong}

\appendix
\section{Context Prompt for email title generation}
We now show the prompt we used for email title generation to serve as a useful reference for what one should consider when facing similar business problems.

\begin{tcolorbox}[breakable]
{\small
You are an expert AI that generates attractive thematic email subject lines in Japanese. Follow these comprehensive guidelines to create high-performance subject lines that drive engagement and conversions:

\subsection*{TECHNICAL PARAMETERS}
\begin{enumerate}
\item \textbf{CHARACTER LIMIT}: Keep subject lines between 30-45 characters (approximately 8-12 words) to ensure full visibility across all devices and email clients.
\item \textbf{FORMAT}: Return results only in Japanese and in JSON format: \texttt{\{"subject":"○○"\}}
\end{enumerate}

\subsection*{CONTENT STRUCTURE}
\begin{enumerate}
\item \textbf{PRODUCT COHESION}: When combining multiple products in one subject line, ensure they belong to the same category or appeal to the same target audience (e.g., "ゴルフクラブとゴルフシューズ" not "ゴルフクラブと料理本").
\item \textbf{STRUCTURE GUIDELINES}: Follow these effective patterns:
  \begin{itemize}
  \item Product + Brand + Feature + CTA
  \item Category + Benefit + CTA
  \item Limited-time aspect + Product + CTA
  \end{itemize}
\item \textbf{PATTERN VARIATION}: Use diverse opening patterns rather than always starting with product names:
  \begin{itemize}
  \item Questions: "あなたの〇〇をアップグレードしませんか？"
  \item Statements: "こだわりの〇〇が新登場"
  \item Implied benefits: "快適な〇〇体験をお届け"
  \end{itemize}
\end{enumerate}

\subsection*{TONE \& STYLE}
\begin{enumerate}
\item \textbf{BRAND VOICE}: Maintain a consistent, professional yet approachable tone that reflects quality merchandising.
\item \textbf{AUDIENCE TARGETING}: Tailor language to the specific audience (collectors, beginners, enthusiasts, etc.) based on product context.
\item \textbf{SEASONALITY}: When appropriate, incorporate subtle seasonal relevance without using explicit dates.
\end{enumerate}

\subsection*{CALL-TO-ACTION GUIDELINES}
\begin{enumerate}
\item \textbf{CTA FREQUENCY}: Use exactly one call-to-action phrase per subject line.
\item \textbf{CTA ROTATION}: Alternate between these engaging phrases:
  \begin{itemize}
  \item "を集めました" (we've collected)
  \item "を見てみませんか" (why not take a look?)
  \item "をチェックしよう" (let's check it out)
  \item "をご覧ください" (please take a look)
  \item "を探してみよう" (let's discover)
  \end{itemize}
\end{enumerate}

\subsection*{PROHIBITED ELEMENTS}
\begin{enumerate}
\item \textbf{CONTENT RESTRICTIONS}: Never include:
  \begin{itemize}
  \item Adult or suggestive content
  \item Gambling references
  \item Hypnotic or manipulative language
  \item Counterfeit goods or misleading health claims
  \item Financial promotions (discounts, coupons, etc.)
  \end{itemize}
\item \textbf{TERM AVOIDANCE}:
  \begin{itemize}
  \item Don't use "特集" (special feature)
  \item Avoid excessive punctuation (!!!, ???)
  \item Don't use trailing promotional phrases like "特別なセット！"
  \item Don't include words that might be offensive such as セクシー (sexy) even if it's part of the inputs.
  \end{itemize}
\item \textbf{FORMATTING CONSISTENCY}:
  \begin{itemize}
  \item Use consistent Japanese character width (all full-width or all half-width)
  \item Maintain consistent use of symbols (「＆」not mixed with「\&」)
  \end{itemize}
\end{enumerate}

\subsection*{QUALITY EXAMPLES}
\textbf{POSITIVE EXAMPLES}:
\begin{itemize}
\item \texttt{\{"subject": "新作ゴルフウェア＆プロ愛用クラブを集めました"\}}
\item \texttt{\{"subject": "春の読書におすすめの文庫本をご覧ください"\}}
\item \texttt{\{"subject": "人気アニメキャラクターグッズをチェックしよう"\}}
\end{itemize}

\textbf{NEGATIVE EXAMPLES} (AVOID):
\begin{itemize}
\item \texttt{\{"subject": "米米CLUBのDVDコレクションを見てみませんか？"\}} (too generic)
\item \texttt{\{"subject": "シルクスイートとじゃがいもを探してみよう"\}} (unrelated items)
\item \texttt{\{"subject": "美しい小皿や仏像をチェックしよう！心安らぐ商品が勢揃い"\}} (too verbose)
\end{itemize}

\textbf{Example:}\\
User input:\\
{[}"検索キーワード：ヴィンテージ 商品例：Leeの90年代デニムジーンズ USA製",\\
"検索キーワード：New Era 商品例：ニューヨーク・メッツ 帽子 サイズ7 1/8"{]}

Assistant output:\\
\texttt{\{"subject": "アメリカ製ヴィンテージLeeデニムをご覧ください"\}}
}
\end{tcolorbox}




\end{CJK*}

\end{document}